\documentclass[runningheads,a4paper]{llncs}

\usepackage{makeidx}  
\usepackage{graphicx}
\usepackage{listings}
\usepackage{booktabs}
\usepackage[font=small,labelfont=bf]{caption}

\usepackage{color}
\usepackage[normalem]{ulem}
\usepackage{subfigure}
\usepackage{cancel}
\usepackage{amssymb}
\usepackage[cmex10]{amsmath}

\newcommand{\ADD}[1]{{{#1}}}

\usepackage{array}
\newcolumntype{L}[1]{>{\raggedright\let\newline\\\arraybackslash\hspace{0pt}}m{#1}}
\newcolumntype{C}[1]{>{\centering\let\newline\\\arraybackslash\hspace{0pt}}m{#1}}
\newcolumntype{R}[1]{>{\raggedleft\let\newline\\\arraybackslash\hspace{0pt}}m{#1}}
\newcommand {\ALGOA}{IUTIS-1} 
\newcommand {\ALGOB}{IUTIS-2}
\newcommand {\ALGOC}{IUTIS-3}
\newcommand {\ALGOD}{IUTIS-5}

\newcommand{\quot}[1]{{``#1''}}

\begin{document}
\frontmatter          
\pagestyle{headings}  
\graphicspath{{images/}}
\mainmatter              
%
\title{How Far Can You Get by Combining Change Detection Algorithms?}

\titlerunning{How Far Can You Get by Combining Change Detection Algorithms?}

\author{Simone Bianco         \and
        Gianluigi Ciocca \and 
        Raimondo Schettini%
}
\authorrunning{S. Bianco, G. Ciocca, R. Schettini}

\institute{
				Department of Informatic Systems and Communications,\\
              University of Milano-Bicocca, Milano, 20126 Italy \\
              \email{\{bianco,ciocca,schettini\}@disco.unimib.it}
}

\maketitle

\begin{abstract}
\ADD{Given the existence of many change detection algorithms, each with its own peculiarities and strengths, we propose a combination strategy, that we termed IUTIS (In Unity There Is Strength), based on a genetic Programming framework. This combination strategy is aimed at leveraging the strengths of the algorithms and compensate for their weakness. In this paper we show our findings in applying the proposed strategy in two different scenarios. The first scenario is purely performance-based. The second scenario performance and efficiency must be balanced. Results demonstrate that starting from simple algorithms we can achieve comparable results with respect to more complex state-of-the-art change detection algorithms, while keeping the computational complexity affordable for real-time applications.}
\keywords{Video surveillance, change detection, algorithm combining and selection, genetic programming, CDNET.}
\end{abstract}

\section{Introduction}
\label{intro}
Many computer vision applications require the detection of changes within video streams, e.g. video surveillance, smart environments, video indexing and retrieval \cite{yu2012,bianco2013cooking-action}. For all these applications, a robust change detection algorithms with a low false alarm rate is required as a pre-processing step. Many algorithms have been proposed to solve the problem of video change detection.
Most of them rely on background subtraction techniques to segment the scene into foreground and background components. \ADD{The outputs of a change detection algorithm are usually binary images of the foreground areas corresponding to moving objects.} \ADD{These algorithms are designed to cope} with the challenges that can be found in a real-world videos such as high variation in environmental conditions, illumination changes, shadows, camera movements and camera-induced distortions and so on. To this end, algorithms are becoming increasingly more complex and thus computationally expensive both in terms of time and memory space. Parallelization of background subtraction algorithms on GPU is a possible way to speed up the computation to make them usable in real-time applications (e.g. \cite{wang2014fast}).

Notwithstanding the improvements, \ADD{it is still difficult to design general-purpose background subtraction algorithms. These algorithms have been demonstrated} to perform well on some types of videos but there is no single algorithm that is able to tackle all the challenges in a robust and computationally efficient way. This can be clearly seen in the CDNET 2014 competition \cite{wang2014cdnet} (ChangeDetection.net) where change detection algorithms are evaluated on a common dataset composed of different types of videos sequences and classified according to their performance. \ADD{To date, more than 35 different algorithms have been evaluated, but many more exists in the literature.}

\ADD{Finally the output of a background subtraction algorithm is usually refined in order to reduce noisy patterns such as isolated pixels, holes, and jagged boundaries.} To improve algorithm accuracy, post-processing of the foreground component, ranging from simple noise removal to more complex object-level techniques, has been investigated. Results indicate that significant improvements in performance are possible if a specific post-processing algorithm is designed and the corresponding parameters are set appropriately \cite{parks2008postproc}.

\ADD{Given the existence of so many change detection algorithms, each which its own peculiarities and strengths, here we are interested in finding how far can we get, in terms of performances, by leveraging existing algorithms to create new ones. Instead of designing from scratch a new algorithm, we combine existing ones with the aim to build a better change detection algorithm. We are interested in testing this idea under two scenarios: a purely performance-based scenario and a performance/efficiency balanced scenario. In \cite{bianco2017tevc} we have investigated the first scenario by considering the best change detection algorithms in the CDNET 2014 competition, disregarding their computational complexity, and combining then under a Genetic Programming (GP) framework \cite{koza1992genetic}. The resulting algorithms significantly outperform the algorithms used in the combination and even other, more recent, approaches. 
In this work, we present our findings with respect to the second scenario. We apply the same general approach used in \cite{bianco2017tevc} but considering state-of-the-art algorithms that are computationally efficient but not top-performing. We want to investigate if also in this scenario, we are able to create an effective algorithm and what kind of performances we can achieve.}

\section{The proposed approach}
\label{sec:gp}
\ADD{In this section, we summarize our GP-based combining approach. A detailed description of the method can be found in \cite{bianco2017tevc}.} GP is a domain-independent evolutionary method that genetically breeds a population of functions, or more generally, computer programs to solve a given problem \cite{koza1992genetic}. \ADD{Evolution is driven by the best fit individuals according to an objective function (i.e. fitness function) that must be maximized or minimized.} The solutions can be represented as trees, lines of code, expressions in prefix or postfix notations, strings of variable length, etc. 
\ADD{GP has been widely used for finding suitable solutions for a wide range of problems needing  optimization. For example, in image processing and computer vision applications GP has been used for: image segmentation, enhancement, layouting, classification, feature extraction, and object recognition \cite{amelio2014evolutionary,bianco2015-user-preferences,corchs2016genetic-programming,sahaf2016automatically,liu2016learning}}.
\ADD{For our purposes,} we feed GP the set of the binary foreground images that correspond to the outputs of the \ADD{single} change detection algorithms, and a set operators represented by unary, binary, and $n$-ary functions that are used to combine the outputs (via logical AND, logical OR, etc\dots) as well as to perform post-processing (via filter operators). 

\ADD{More formally,} given a set of $n$ change detection algorithms $\mathcal{C}=\{C_i\}_{i=1}^n$, the solutions evolved by GP are built using the set of functionals symbols $\mathcal{F}$ and the set of terminal symbols $\mathcal{T}=\mathcal{C}$. 
We build the set of functionals symbols considering operators that work in the spatial neighborhood of the image pixel, or combine (stack) the information at the same pixel location but across different change detection algorithms. The list of functional symbols used is given below:
\ADD{
\begin{itemize}
\item[-]ERO (Erosion): it requires one input, works in the spatial domain and performs morphological erosion with a $3\times3$ square structuring element;
\item[-]DIL (Dilation): it requires one input, works in the spatial domain and performs morphological dilation with a $3\times3$ square structuring element;
\item[-]MF (Median Filter): it requires one input, works in the spatial domain an performs median filtering with a $5 \times 5$ kernel;
\item[-]OR (Logical OR): it requires two inputs, works in the stack domain and performs the logical $OR$ operation;
\item[-]AND (Logical AND): it requires two inputs, works in the stack domain and performs the logical $AND$ operation;
\item[-]MV (Majority Vote): it requires two or more inputs, works in the stack domain and performs the majority vote operation;
\end{itemize}
}

We define the fitness function used in GP by taking inspiration from the CDNET website, where change detection algorithms are evaluated using different performance measures and ranked accordingly. 
Given a set of video sequences $\mathcal{V}=\{V_1,\ldots,V_S\}$, a set of performance measures $\mathcal{M}=\{m_1,\ldots,m_M\}$ the fitness function of a candidate solution $C_0$, $f(C_0)$ is defined as the average rank across video sequences and performance measures:

\begin{equation}
f(C_0)= \frac{1}{M} \sum_{j=1}^M  \bigg( {\rm{rank}}_{C_0} \Big( C_0; \big\{ m_j\big(C_k(\mathcal{V})\big) \big\} _{k=1}^{n} \Big)  +  
   \sum_{i=1}^2{w_i P_i(C_0)} \bigg)
\label{eq:fitness}
\end{equation}

\noindent
where ${\rm{rank}_{C_0}}(\cdot)$ computes the rank of the candidate solution $C_0$ with respect to the set of algorithms $\mathcal{C}$ according to the measure $m_j$. $P_1(C_0)$ is defined as the signed distance between the candidate solution $C_0$ and the best algorithm in $\mathcal{C}$ according to the measure $m_j$:

\begin{equation}
P_1(C_0)=
\begin{cases}
\displaystyle
-m_j(C_0(\mathcal{V}))+\max_{C_k \in \mathcal{C}}{m_j(C_k(\mathcal{V}))} \\
\qquad \qquad \qquad \qquad \mbox{if the higher $m_j$ the better} \\
\displaystyle
 m_j(C_0(\mathcal{V}))-\min_{C_k \in \mathcal{C}}{m_j(C_k(\mathcal{V}))} \\
\qquad \qquad \qquad \qquad \mbox{if the lower  $m_j$ the better} \\
\end{cases}
\label{eq:penalty1}
\end{equation}
\noindent
and $P_2(C_0)$ is a penalty term corresponding to the number of different algorithms selected for the candidate solution $C_0$:

\begin{equation}
P_2(C_0) = \frac{\mbox{\# of algorithms selected in } C_0}{\mbox{\# of algorithms in } \mathcal{C}}
\label{eq:penalty2}
\end{equation}

The role of $P_1$ is to produce a fitness function $f(C_0) \in \mathbb{R}$, so that in case of candidate solutions having the same average rank, the one having better performance measures is considered a fitter individual in GP. The penalty term $P_2$ is used to force GP to select a small number of algorithms in $\mathcal{C}$ to build the candidate solutions. The relative importance of $P_1$ and $P_2$ is independently regulated by the weights $w_1$ and $w_2$ respectively. 

\section{Experimental setup}
\label{sec:setup}

\ADD{Since we wanted to test computationally efficient and simple change detection algorithms, we chose the set of change detection algorithms} $\mathcal{C}$ to be combined among those implemented in the BGSLibrary\footnote{\url{https://github.com/andrewssobral/bgslibrary}}. BGSLibrary is a free, open source and platform independent library which provides a C++ framework to perform background subtraction using code provided by different authors. We used the 1.9.1 version of the library which implements more than 30 different algorithms. 
We base our choice of the algorithms on the recent review paper of the authors of BGSLibrary \cite{Sobral2014} where the computational costs as well as the performances of the different algorithms have been assessed. The rationale is to use computationally efficient algorithms having above average performances, and possibly exploiting different background subtraction strategies. Based on the results in \cite{Sobral2014}, and on some preliminary tests that we have performed, we selected the following algorithms: 
\ADD{
Static Frame Difference (SFD), 
Adaptive-Selective Background Learning (ASB), 
Adaptive Median (AM)\cite{DPAdaptiveMedian}, 
Gaussian Average (GA)\cite{DPWrenGA}, 
Gaussian Mixture Model (ZMM)\cite{DPZivkovicAGMM06}, 
Gaussian Mixture Model (MoG)\cite{LBMixtureOfGaussians}, 
Gaussian Mixture Model (GMM)\cite{DPGrimsonGMM}, 
Eigenbackground/ SL-PCA (EIG)\cite{DPEigenbackground}, 
VuMeter (VM)\cite{VuMeter}, 
$\Sigma\Delta$ Background Estimation (SD)\cite{SigmaDelta09}, 
Multiple Cues (MC)\cite{SJNMultiCue}. 
}
All the algorithms have been tested in \cite{Sobral2014} with the exception of SigmaDelta and SJNMultiCue algorithms. These have been added in recent versions of the BGSLibrary. We decide to include them since they show interesting performances although they are slightly more computationally intensive with respect to the simpler algorithms.

The performance measures $\mathcal{M}$ are computed using the framework of the CDNET 2014 challenge \cite{wang2014cdnet}. The framework implements the following seven different measures: recall, specificity, false positive ratio (FPR), false negative ratio (FNR), percentage of wrong classifications (PWC), precision, and F-measure. A ranking of the tested algorithms is also computed starting from the partial ranks on these measures.
The CDNET 2014 dataset is composed of 11 video categories, with four to six videos sequences in each category. The categories exhibit different video contents and are chosen to test the background subtraction algorithms under different operating conditions. The challenge rules impose that each algorithm should use only a single set of parameters for all the video sequences. For this reason we set the parameters of the algorithms to their default values, i.e. the values in the configuration files provided in the BGSLibrary. 

\ADD{We set the parameters of GP as in \cite{bianco2017tevc}. Also, the GP solutions are generated by considering the shortest video sequence in each of the 11 CDNET 2014 categories as training set. The images in this set are less than 10\% of the total images in the whole dataset; this minimizes the over-fitting effect if more images were used. We name the best solution found by GP in this way as \ALGOB{ } (the term IUTIS is derived by quoting the Greek fabulists Aesop 620BC-560BC: \quot{In Unity There Is Strength}). We also created a different algorithm, \ALGOA, by considering a smaller training set composed of all video sequences in the \quot{Baseline} category. As the name suggests, this category contains basic video sequences. \ALGOA{ } exhibits worse performances than \ALGOB, and since its results are not directly comparable with the reported ones, it will not be further considered in the discussion.}

\section{Results}
\label{sec:results}

\begin{figure*}[!tbp]
\resizebox{\linewidth}{!}{
\begin{tabular}{cc}
\includegraphics[width=\columnwidth]{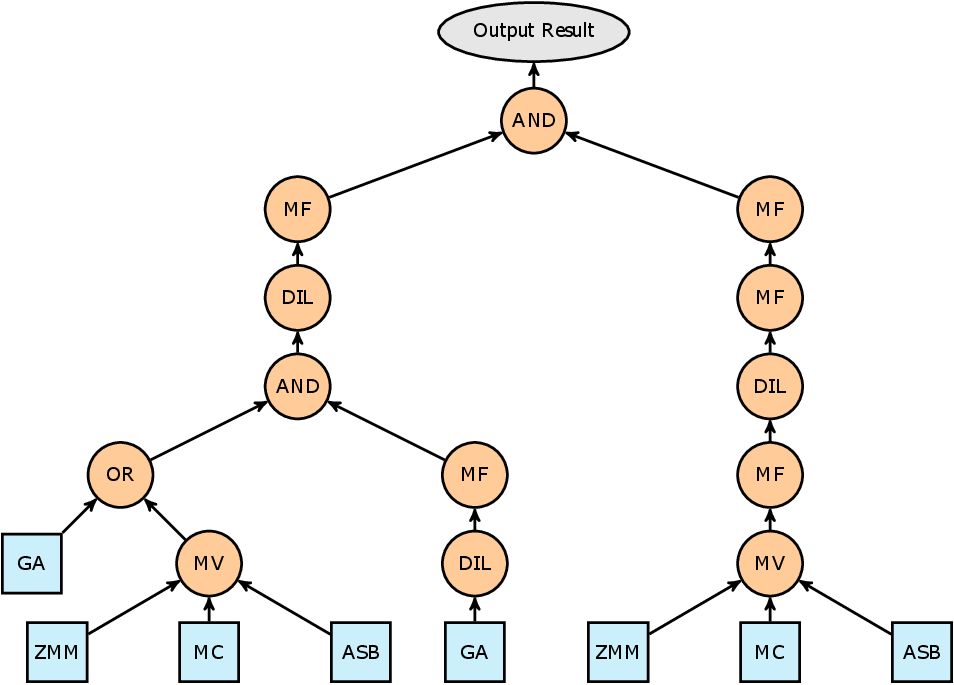} &
\includegraphics[width=\columnwidth]{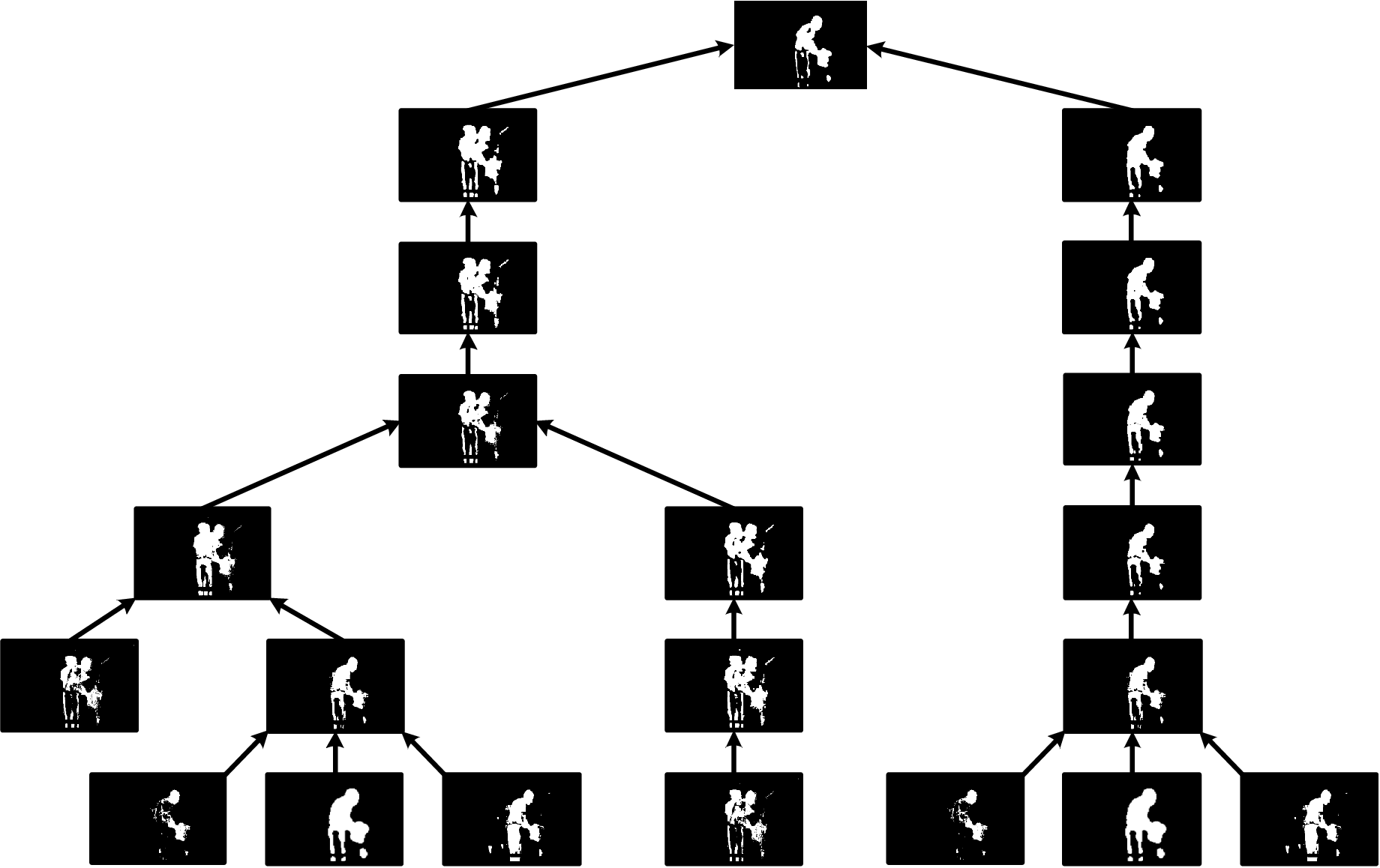}\\
\end{tabular}
}
\caption{\ALGOB{ } solution tree and example masks.}
\label{fig:solNUOVA-tree-masks}
\end{figure*}

\begin{table*}[!tbp]
\setlength\extrarowheight{3pt}
\caption{Detailed evaluation results of the \ALGOB{} algorithm for each category of the evaluation dataset.}
\label{tab:iutis2}
\begin{center}
\resizebox{\textwidth}{!}{ 
\begin{tabular}{lp{1.4truecm}p{1.5truecm}p{1.4truecm}p{1.4truecm}p{1.4truecm}p{1.4truecm}p{1.5truecm}}
\hline
Scenarios & Recall	&	Specificity	&	FPR		&		FNR		&		PWC		&		Precision	&	F-measure \\
\hline			
Overall & 0.6703 & 0.9846 & 0.0154 & 0.3297 & 2.9932 & 0.7191 & 0.6163\\
Bad Weather & 0.6388 & 0.9995 & 0.0005 & 0.3612 & 0.6290 & 0.9380 & 0.7525\\
Low Framerate & 0.7295 & 0.9947 & 0.0053 & 0.2705 & 1.2079 & 0.7160 & 0.6395\\
Night Videos & 0.6089 & 0.9861 & 0.0139 & 0.3911 & 2.2366 & 0.4706 & 0.5157\\
PTZ & 0.8329 & 0.8939 & 0.1061 & 0.1671 & 10.6978 & 0.1884 & 0.2397\\
Turbulence & 0.8444 & 0.9988 & 0.0012 & 0.1556 & 0.2349 & 0.7737 & 0.7967\\
Baseline & 0.7452 & 0.9978 & 0.0022 & 0.2548 & 1.5115 & 0.9100 & 0.7913\\
Dynamic Background & 0.8027 & 0.9828 & 0.0172 & 0.1973 & 2.0051 & 0.5564 & 0.5741\\
Camera Jitter & 0.7209 & 0.9867 & 0.0133 & 0.2791 & 2.4236 & 0.7184 & 0.7165\\
Intermittent Object Motion & 0.3735 & 0.9973 & 0.0027 & 0.6265 & 4.7669 & 0.8374 & 0.4836\\
Shadow & 0.6636 & 0.9946 & 0.0054 & 0.3364 & 2.2199 & 0.8621 & 0.7393\\
Thermal & 0.4125 & 0.9987 & 0.0013 & 0.5875 & 4.9923 & 0.9395 & 0.5306 \\ \hline
\end{tabular}
}
\end{center}
\end{table*}

The tree structure for the \ALGOB{} solution is shown in Figure \ref{fig:solNUOVA-tree-masks}.
\ADD{In the same figure an example of the output at each node on a sample frame is also reported.
From the solution tree it is possible to notice that \ALGOB{} selected and combined a subset of four simple change detection algorithms out of the 11 available: it selected GA, ZMM, MC, and ASB. Concerning the tree structure, we can notice that \ALGOB{} presents a single long branch in its right-hand side. Starting from the functionals defined in Section \ref{sec:gp}, GP was able to create new ones. For instance, the solution tree uses a sequence of the operator $\text{MF}$, which can be seen as an approximation of what could be obtained using a larger kernel for the median filter. 
}
The detailed results of the 
\ALGOB{} algorithm, computed using the evaluation framework of the CDNET 2014 challenge on its 11 video categories, are reported in Table 
\ref{tab:iutis2}.
\ADD{
The overall F-measure of \ALGOB{} and of all the change detection algorithms considered are reported in Table \ref{tab:tabellafmeasures1}.}
\ADD{From the values reported it is possible to see that our solution is better than the best algorithm fed to GP (i.e. MC), achieving a F-measure that is 7.2\% higher.}

%
%


\begin{table}[!hbp]
\centering
\caption{\ADD{Overall F-measure all the change detection algorithms considered to build \ALGOB{} (left), and \ALGOC{}, \ALGOD{} and IUTIS-7 (right). An empty circle means that the algorithm was in $\mathcal{C}$ but was not selected, a full circle otherwise.}}
\label{tab:tabellafmeasures1}
\resizebox{0.97\textwidth}{!}{
\begin{tabular}{cc}
\begin{tabular}{lccc}
\hline
Method  & F-meas. & Used by & Impr. by\\
        & 		  & IUTIS-2 & IUTIS-2\\
\hline
ASB								&		0.4501 & $\bullet$  & 0.1662\\
ZMM	\cite{DPZivkovicAGMM06}		&		0.5175 & $\bullet$	& 0.0988\\
GA	\cite{DPWrenGA}				&		0.4535 & $\bullet$	& 0.1628\\
MC	\cite{SJNMultiCue}			&		0.5444 & $\bullet$	& 0.0719\\
SFD								&		0.2626 & $\circ$			& 0.3537\\
AM \cite{DPAdaptiveMedian}		&		0.4029 & $\circ$			& 0.2134\\
GMM	\cite{DPGrimsonGMM}			&		0.4589 & $\circ$			& 0.1574\\
EIG	\cite{DPEigenbackground}	&		0.3215 & $\circ$			& 0.2948\\
MoG	\cite{LBMixtureOfGaussians}	&		0.4304 & $\circ$			& 0.1859\\
VM	\cite{VuMeter}				&		0.3990 & $\circ$			& 0.2173\\
SD	\cite{SigmaDelta09}			&		0.3969 & $\circ$			& 0.2194\\
\hline
MV-11							& 		0.5098 & 			& 0.1065\\
IUTIS-2 						&		\bf{0.6163} &  		& -.----\\
\hline
& & & \\
& & & \\
\end{tabular}

& \quad

\begin{tabular}{lccrc}
\hline
Method  & F-meas. & Used by   & Impr. by  & Impr. by\\
        &         & IUTIS-3/5/7 & IUTIS-3	  & IUTIS-5\\
\hline
FTS \cite{wang2014static}		& 0.7281 &  $\bullet / \bullet / \bullet$& 0.0413  &	0.0540\\
SBS	\cite{subsense}				& 0.7092 &  $\bullet / \bullet / \bullet$& 0.0602  &	0.0729\\
CWS	\cite{gregorio2014change}	& 0.7050 &	$\bullet / \bullet / \bullet$& 0.0644  &	0.0771\\
SPC	\cite{sedky2014spectral}	& 0.6932 &	$ \phantom{\bullet} / \bullet / \bullet$  & 0.0762  &	0.0889\\
AMB	\cite{wang2014fast}			& 0.7058 &	$ \phantom{\bullet}/ \bullet / \bullet$  & 0.0636  &	0.0763\\
KNN \cite{DPZivkovicAGMM06}		& 0.5984 &	$ \phantom{\bullet} / \phantom{\bullet} / \,\, \circ$	   & 0.1710  &	0.1837\\
SCS	\cite{maddalena2012sobs}	& 0.6572 &	$ \phantom{\bullet} / \phantom{\bullet} / \,\, \circ$	   & 0.1122  &	0.1249\\
RMG	\cite{varadarajan2013spatial}& 0.6282 &		   & 0.1412  &	0.1539\\
KDE	\cite{elgammal2000non}		& 0.5689 &		   & 0.2005  &	0.2132\\
\hline
MV-3							& 0.7496 &	  	     & 0.0198 & 			 	0.0325\\
MV-5							& 0.7569 &	         & 0.0125 & 			 	0.0252\\
MV-7							& 0.7115 &	         & 0.0579 & 			 	0.0706\\
IUTIS-3							& 0.7694 &	         & -.---- & 			 	0.0127\\
IUTIS-5							& \bf{0.7821} &	     &-0.0127 & 			 	-.----\\
IUTIS-7							& \bf{0.7821} &	     &-0.0127 & 			 	0.0000\\
\hline
\end{tabular}\\
\end{tabular}
}
\end{table}

\begin{figure}[!tbp]
\centering
\begin{tabular}{cp{2em}c}
\includegraphics[width=0.35\columnwidth]{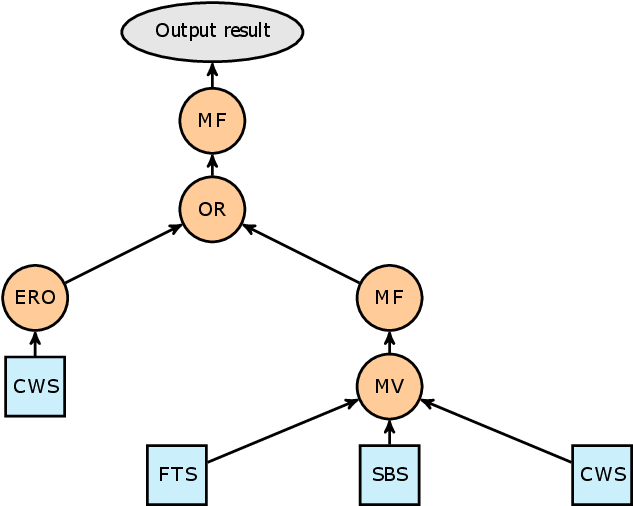} & &
\includegraphics[width=0.5\columnwidth]{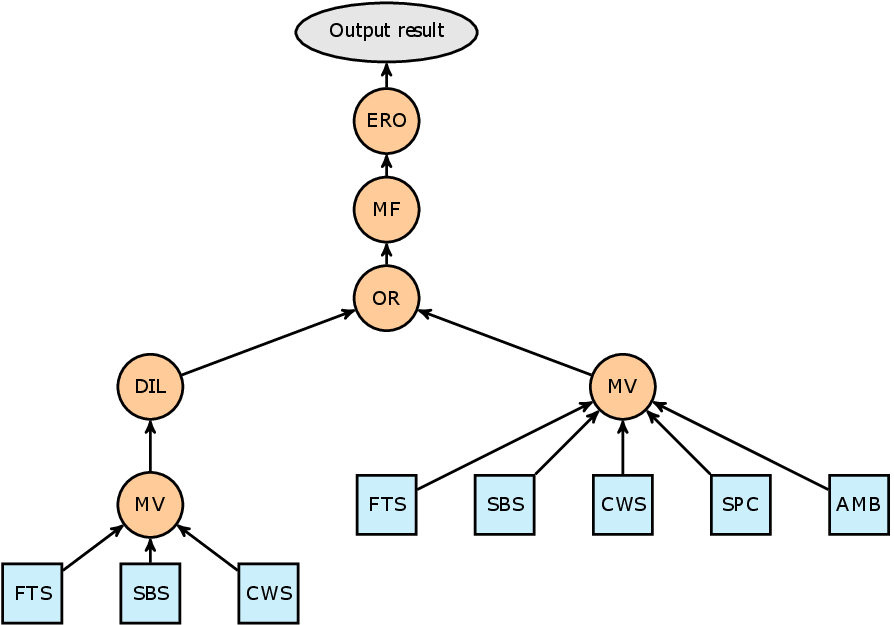}\\
\end{tabular}
\caption{\ALGOC{ } solution tree (left) and \ALGOD{ } solution tree (right).}
\label{fig:solOTHERS-tree}
\end{figure}

\ADD{For comparison, we also report here (see Figure \ref{fig:solOTHERS-tree}) the solution trees of \ALGOC{} and \ALGOD{}, that we recall have been generated by the same method here described but in a purely performance-based scenario \cite{bianco2017tevc}.
The set of algorithms $\mathcal{C}$ available for GP to build \ALGOC{} were the three top performing algorithms on CDNET 2014, i.e.: Flux Tensor with Split Gaussian models (FTS)\cite{wang2014static}, Self-Balanced SENsitivity SEgmenter (SBS)\cite{subsense}, and Change Detection with Weightless Neural Networks (CWS) \cite{gregorio2014change}. The set $\mathcal{C}$ available for \ALGOD{} also included Change Detection based on Spectral Reflectaces (SPC) \cite{sedky2014spectral} and Extension of the Adapting Multi-resolution Background Extractor (AMB)\cite{wang2014fast}. \\ From the comparison of the solution trees reported in Figure \ref{fig:solNUOVA-tree-masks} and \ref{fig:solOTHERS-tree}, it is possible to notice that \ALGOB{} is more complex in terms of functionals used (see for example the right branch). This is due to the fact that very simple algorithms are used and more operations are needed on their output to achieve higher performance. The overall F-measure of \ALGOC{}, \ALGOD{} and of all the change detection algorithms considered are reported in Table \ref{tab:tabellafmeasures1}. From the results it is possible to see that \ALGOC{} is better than any other single algorithms, with a F-measure that is 4.13\% higher than the best algorithm used by GP. In the case of \ALGOD{} this difference increases to 5.4\%. It is worth noting that all our solutions are better than majority vote solutions (denoted with MV) applied to the corresponding sets $\mathcal{C}$. In \cite{bianco2017tevc} we also experimented  with larger cardinalities of $\mathcal{C}$, i.e. $\#\mathcal{C}=7$ and 9, but in both cases the corresponding solutions found by GP, i.e. IUTIS-7 and IUTIS-9, obtained identical performance with respect to \ALGOD{} and thus we only report them in Table \ref{tab:tabellafmeasures1}.}

\begin{figure*}
\resizebox{\textwidth}{!}{ 
\centering
\includegraphics{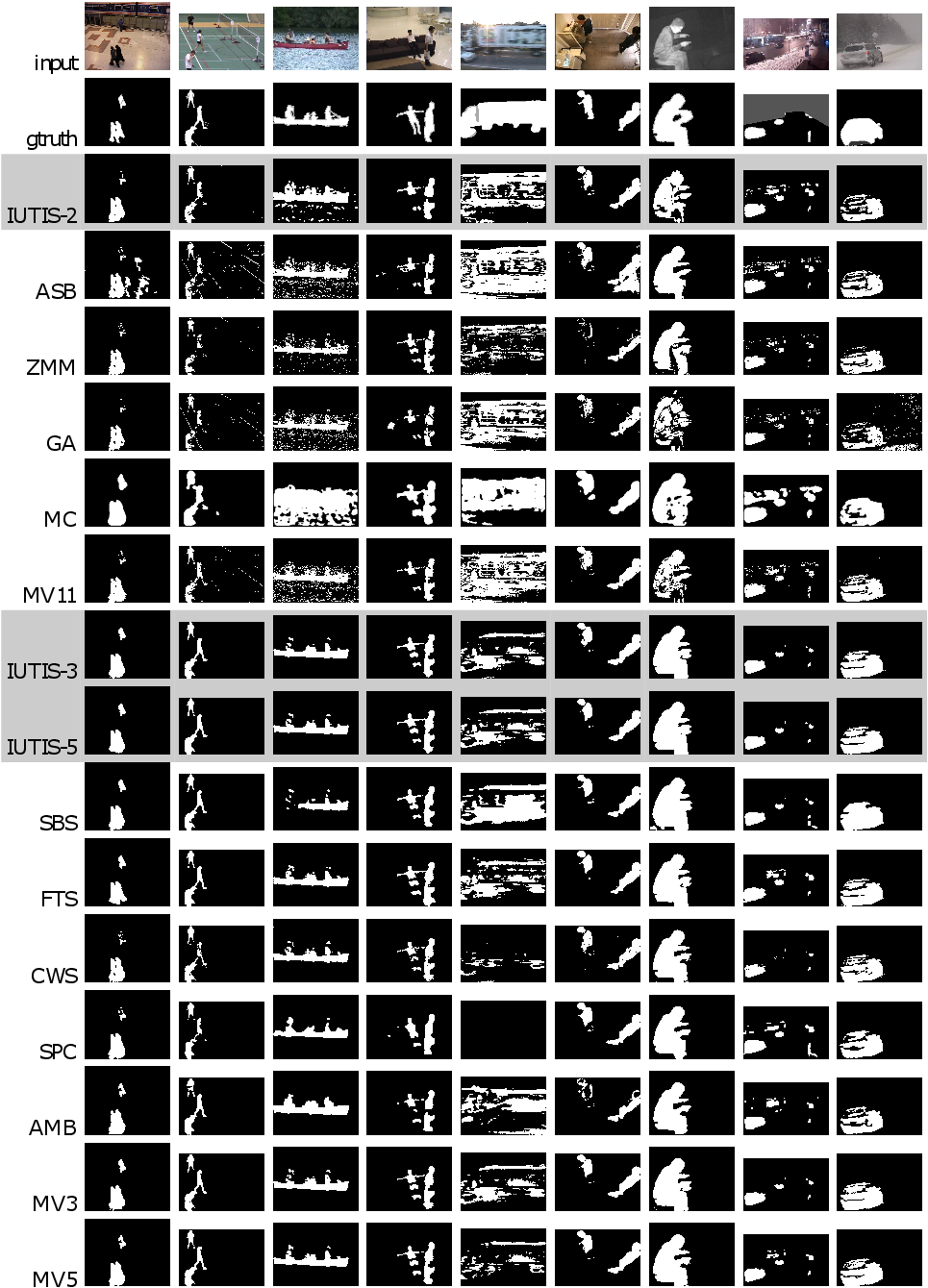}
}
\caption{Examples of binary masks created by the proposed algorithms and some of the algorithms used in the combination.}
\label{fig:esempi}
\end{figure*}

Outputs of some of the tested algorithms on sample frames in the CDNET 2014 dataset, together with input images and ground truth masks, are shown in Figure \ref{fig:esempi}. 

\section{Computational Time}
\label{sec:computational}
\ADD{\ALGOB{} algorithm has been implemented in C++ and use the OpenCV library for image processing. Table \ref{tab:time} reports the computational time of the proposed algorithm in frames per seconds. For evaluation purpose, we have implemented two versions of \ALGOB: a \quot{Sequential} one, and a \quot{Parallel} one. The first version refers to the implementation of the algorithms without any particular optimization. While the second one refers to an optimized implementation of the algorithms obtained by exploiting parallelism on a multicore CPU. We used the OpenMP directives (\texttt{parallel for} and \texttt{sections}) to parallelize both the computation of the masks, and the execution branches of the solution tree. The timing measurements are carried out on a 3.3Ghz Intel Core-i5 (quadcore) with 16GB RAM and Windows 7 Professional operating system.
As it can be seen, the \ALGOB{ } algorithm can be efficiently parallelized. Specifically, the frame rates of the parallel version is, on average, about 2 times faster than that the sequential version.

In table \ref{tab:time} we also report the computational time of \ALGOC{ }, and \ALGOD used in \cite{bianco2017tevc}. For these algorithms, the computational time is an estimated of an hypothetical parallel implementation and corresponds to the slowest algorithm used in the solution tree. For completeness we compare the computational time of the different IUTIS algorithms with the top five algorithms in Table \ref{tab:tabellafmeasures1} (right). The slowest algorithm, with 10 frame-per-seconds, is FTS (that is also used in \ALGOC{ } and \ALGOD). This algorithm is implemented in MATLAB while the other algorithms are all implemented in C++. The AMB algorithm is the most efficient one with an impressive 843 frame per second. This result is achieved thanks to the parallel implementation on GPU using the CUDA architecture.
}

\begin{table}[!tbp]
\caption{Computational time, in frames per seconds and at the resolution of $320\times240$ pixels, of different change detection algorithms on a i5-2500K@3.3Ghz computer with 16 GB RAM. For \ALGOC{ }and \ALGOD{ }we report an estimate corresponding to the slowest algorithm in an hypothetical parallel implementation.} 
\label{tab:time}
\setlength\extrarowheight{3pt}
\setlength{\tabcolsep}{12pt}
\center
\resizebox{0.60\textwidth}{!}{ 
\begin{tabular}{llc}
\hline
Algorithm & Implementation & FPS@320x240 \\ \hline
\ALGOB & C++, Sequential & 18 \\ 
\ALGOB & C++, Parallel & 40 \\ 
\ALGOC & Misc & 10 (Parallel Estimate) \\
\ALGOD & Misc & 10 (Parallel Estimate) \\  \hline
FTS & MATLAB & 10 \\  
SBS & C++ & 31 \\ 
CWS & C++, OpenMP & 18 \\ 
SPC & C++ & 12 \\ 
AMB & C++ , CUDA & 843 \\ \hline
\end{tabular}
}
\end{table}

\section{Conclusion}
\label{sec:conclusion}
In this paper we have presented an evolutionary approach, based on Genetic Programming, to combine simple change detection algorithms to create a more robust algorithm. 
The solutions provided by Genetic Programming allow us to select a subset of the simple algorithms. Moreover, we are able to automatically combine them in different ways, and perform post-processing on their outputs using suitable operators to produce the best results.
\ADD{Our combination strategy, is able to produce algorithms that are more effective in solving the change detection problem in different scenario. If we are interested in obtaining the maximum performance disregarding the computational complexity of the algorithms themselves, we can combining few top-performing algorithms and achieve the best overall performances (i.e. \ALGOC{} and \ALGOD{}). On the contrary, if we want to improve the performances of existing algorithms while maintaining a limited computational complexity, we can effectively combine several simple algorithms and} achieve comparable results of more complex state-of-the-art change detection algorithms (i.e. \ALGOB). In particular, the parallelized version of \ALGOB{} exhibits remarkable performance while being computationally affordable for real-time applications.


\bibliographystyle{splncs03}   


%
%

\end{document}